\pdfoutput=1
\documentclass[11pt]{article}

\usepackage{naacl2021}

\usepackage{times}
\usepackage{latexsym}

\usepackage[T1]{fontenc}
\usepackage[utf8]{inputenc}

\usepackage{microtype}

\usepackage{graphicx}

\title{Extracting Similar Questions From Naturally-occurring Business Conversations}

\author{Xiliang Zhu \and David Rossouw \and Shayna Gardiner \and Simon Corston-Oliver \\
        Dialpad Canada Inc.\\ 1100 Melville St \#400\\ 
        Vancouver, BC, Canada, V6E 4A6\\
        \texttt{\{xzhu, davidr, sgardiner, scorston-oliver\}@dialpad.com} \\}

\begin{document}
\maketitle
\begin{abstract}
Pre-trained contextualized embedding models such as BERT are a standard building block in many natural language processing systems. We demonstrate that the sentence-level representations produced by some off-the-shelf contextualized embedding models have a narrow distribution in the embedding space, and thus perform poorly for the task of identifying semantically similar questions in real-world English business conversations. We describe a method that uses appropriately tuned representations and a small set of exemplars to group questions of interest to business users in a visualization that can be used for data exploration or employee coaching.
\end{abstract}

\section{Introduction}

We describe a real-world application that identifies semantically similar questions in English language business telephone calls between a customer and a sales or support agent. Identifying questions with similar meanings but different formulations provides insight into customer conversations that can be used for business analytics or for training agents on how to respond better to customer needs.

Pre-trained contextualized embeddings such as those that can be extracted from BERT \citep{devlin-etal-2019-bert} and GPT-2 \citep{radford2019language} have become essential building blocks for natural language processing (NLP) solutions. Embeddings extracted from these models can be used directly in unsupervised NLP applications, such as clustering and similarity comparison. Starting with pre-trained models reduces the need for expensive training and  is particularly important in real-world settings, where high-quality data annotation is a major challenge. 

At our company we offer transcription of calls via automatic speech recognition and natural language understanding services for business conversations that enable product features such as sentiment analysis, identification of action items, and note-taking.

Questions play a key role in natural conversation. It would be of great value to users if we could group questions asked in the calls that are semantically similar, giving our clients deeper insight in what types of questions their customers are asking. It is worth noting here that questions that arise naturally in the course of human-to-human conversation are quite different from the reading comprehension questions in public question answering datasets such as Stanford Question Answering Dataset (SQuAD) \citep{rajpurkar-etal-2016-squad}, making this problem all the more challenging.  We provide a detailed comparison in section 4.1.

In this work, we demonstrate that the representations of question sentences only occupy a narrow cone in the output embedding space, a property known as \textit{anisotropy} \citep{ethayarajh-2019-contextual}. This property makes it difficult to directly use the embeddings as input to semantic comparison functions. Then, we show how  Sentence-BERT \citep{reimers-gurevych-2019-sentence}, a fine-tuned model based on BERT can overcome this problem and improve the performance in our applied task. Additionally, we propose an exemplar-based semantic matching approach to efficiently find sentences that are semantically similar to a specific topic.

\begin{table*}[t]
\centering

\begin{tabular}{| p{2cm} | p{3cm}| p{3cm} | p{3cm}| p{3cm}|}
\textbf{Topic} &
\textbf{Exemplar 1} &
\textbf{Exemplar 2} &
\textbf{Exemplar 3} &
\textbf{Exemplar 4} \\
\hline
Pricing &
How much is it monthly? &
What’s the lowest price? &
What is this like in terms of pricing? &
How much do you charge?\\
\hline
Contact Information &
Can I get your name? &
Can you give me your email address? &
What is your last name? &
Would you mind giving me the phone number? \\

\end{tabular}
%}
\caption{Selected exemplar questions for topics “Pricing” and “Contact Information”}
\label{table1}
\end{table*}

\section{Background}
\subsection{Pre-trained Contextualized Embeddings}
There have been various pre-trained contextualized models based on deep neural networks in recent years. They include: 
\begin{itemize}
\item BERT \citep{devlin-etal-2019-bert}: one of the most popular large pre-trained models. Pooling over the last hidden state of BERT can be used as the sentence representation in many applications. Additionally, BERT prepends a \textit{CLS} token (short for ``classification'') to the start of each input sentence, which is trained by the Next Sentence Prediction task to represent sentence-level classification.

\item GPT-2 \citep{radford2019language}: A large pre-trained language model that shows superior results in many NLP tasks. We use the pooling of the last hidden state as the representation of the sentence.

\item Sentence-BERT (SBERT) \citep{reimers-gurevych-2019-sentence}: A modification of pre-trained BERT that is fine-tuned on Natural Language Inference (NLI) \citep{bowman-etal-2015-large} and Semantic Textual Similarity (STS) \citep{cer-etal-2017-semeval} datasets. SBERT achieves better performance than BERT in semantic textual similarity benchmarks.

\end{itemize}

\subsection{Industry Applications}
At our company, we provide VoIP (Voice over Internet Protocol) telephony for businesses, with products for sales and support call centres. Calls are optionally recorded and transcribed using an in-house Automatic Speech Recognition (ASR) system. In addition to transcription, we offer NLP analysis of our in-house transcripts, e.g., identifying and extracting questions asked by customers in calls. 

We group extracted questions into semantically related sets. Questions within a set may vary in their syntactic structure  but have similar or equivalent meaning. For instance, people can ask questions in various forms when inquiring about pricing, such as “What does it cost?”, “How much is it?” or “What’s the price of that?” 

Grouping related questions can help managers better coach their agents and prepare answers for commonly asked questions. Since hand-labelling our call transcripts would be expensive and time-consuming, we leveraged the rich semantic and syntactic information encoded in contextualized embeddings to build a semantic matching pipeline that requires no supervision or data labelling.

\section{Methods}
\subsection{Sentence Representation}
We experiment with several contextualized embedding models, using these models to construct sentence representations for our question dataset. For each model, we also test different methods for extracting embeddings as the sentence representations. The models and representations in our experiments include:

\begin{itemize}
\item BERT base: we compare using the \textit{CLS} token versus average pooling of the last hidden layer, with an embedding size of 768, noted as “BERT-base-cls” and “BERT-base-mean'' respectively.

\item GPT-2: average pooling of last hidden state with a size of 768, referred to as “GPT2-mean”.

\item SBERT base fine-tuned on the NLI and STS datasets: we compare using the \textit{CLS} token and average pooling of the last hidden layer with an embedding size of 768. These are referred to as “SBERT-base-cls” and “SBERT-base-mean'' respectively.

\end{itemize}

\begin{table*}[t]
\centering
\begin{tabular}{| c| c| c | c|}
 &
\textbf{Average question length} &
\textbf{Median question length} &
\textbf{Percentage of stop words} \\
Our data&
8.12 &
7 &
64.4\% \\
SQuAD&
10.20 &
10 &
46.0\% \\
\end{tabular}
%}
\caption{Comparison between questions in our data and SQuAD, each with 10,000 samples}
\label{table2}
\end{table*}

\subsection{Exemplar-based Semantic Matching}
We propose an efficient semantic matching formula that leverages the power of sentence representations from pre-trained models, and uses them directly to find semantically similar sentences for predefined semantic catgeories. 

Based on discussions with customers, we identified frequently occurring topics that provide business insight. For each predefined question topic, we manually identify a few exemplar sentences exhibiting a variety of lexical and syntactic formulations to use as the references for the semantic matching. Table 1 gives selected exemplars for two business topics, “Pricing” and “Contact Information”.

We compare two formulae for measuring semantic similarity given the sentence embeddings. The first formula is an unweighted similarity score of any query against the topic defined as:

\begin{equation}
\label{eq:1}
score = \frac{\sum_{i=1}^{N} \cos (q, s_i)}{N}
\end{equation}

\(N\) is the number of exemplar sentences, \(q\) denotes the embedding representation of the query sentence, while \(s_i\) is the embedding representation of the \(i^{th}\) exemplar sentence. Cosine similarity is used as the measurement of similarity as it is the most commonly used directional metric in vector space. We pair the query sentence with each exemplar question and compute the cosine similarity score, then average over the number of exemplars.

We compare the unweighted score to a weighted similarity score defined as:
\begin{equation}
\label{eq:2}
score = \frac{\sum_{i=1}^{N} p_i}{N}
\end{equation}
where if \(\cos (q, s_i) < threshold\):
\[
p_i=\cos (q, s_i)
\]
else:
\[
p_i=w\cdot N
\]

When measuring a match between a given new question with exemplars, the unweighted score treats all exemplar questions equally, whereas the weighted score emphasizes matches to specific exemplars.

The \(threshold\) is a manually chosen parameter based on the \(90^{th}\) percentile of the cosine similarity score distribution over all pairs in our semantic matching candidate pool. The parameter \(w\) is a manually selected constant (typically in the range \( w  \epsilon [3, 5]\) in our application) that biases towards highly similar pairs.

\begin{figure*}[t]
\centering
\includegraphics[width=0.63\textwidth]{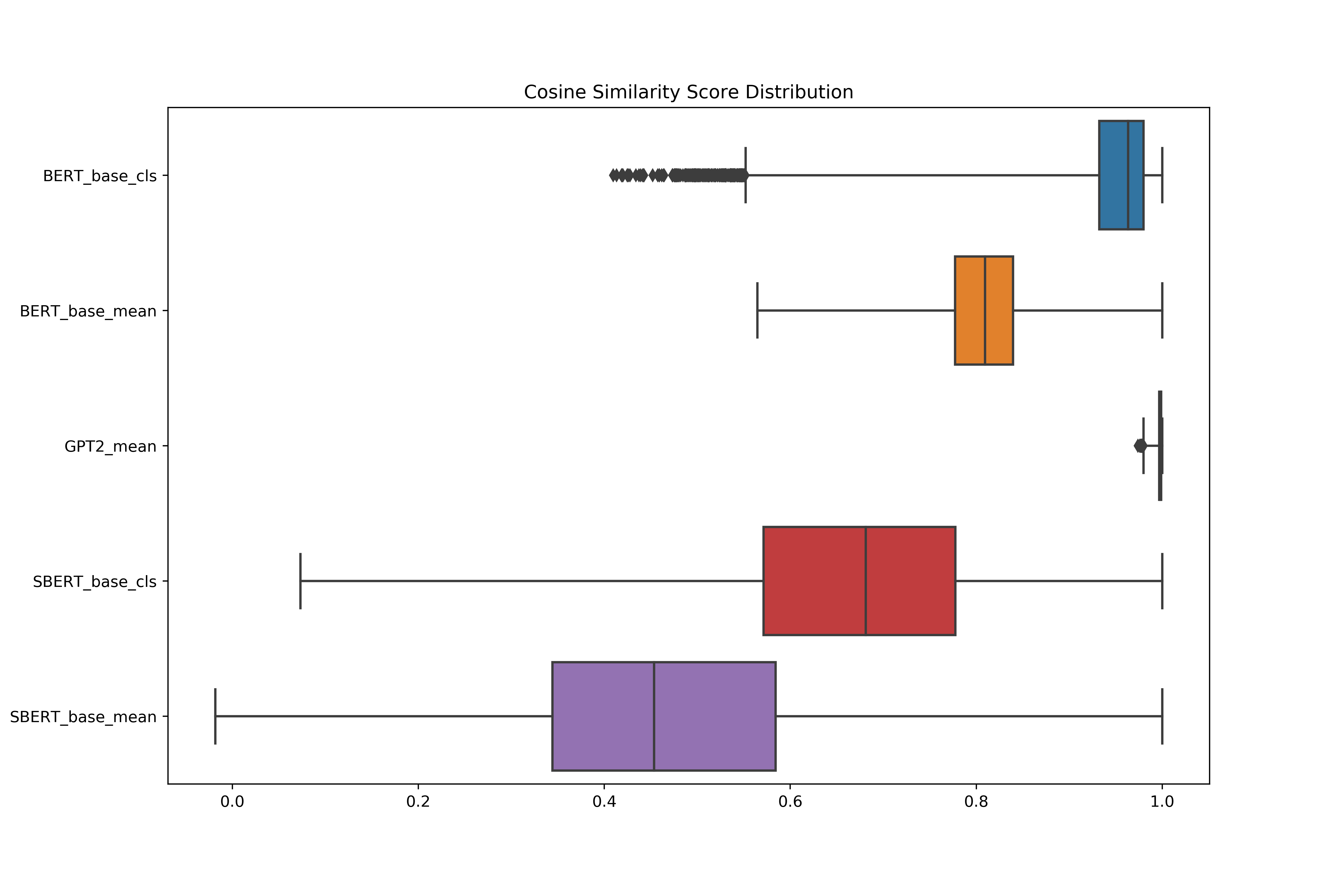} % Reduce the figure size so that it is slightly narrower than the column.
\caption{The cosine similarity score distribution of sentence embedding vectors from different models. }
\label{fig1}
\end{figure*}

\begin{figure*}[t]
\centering
\includegraphics[width=0.3\textwidth]{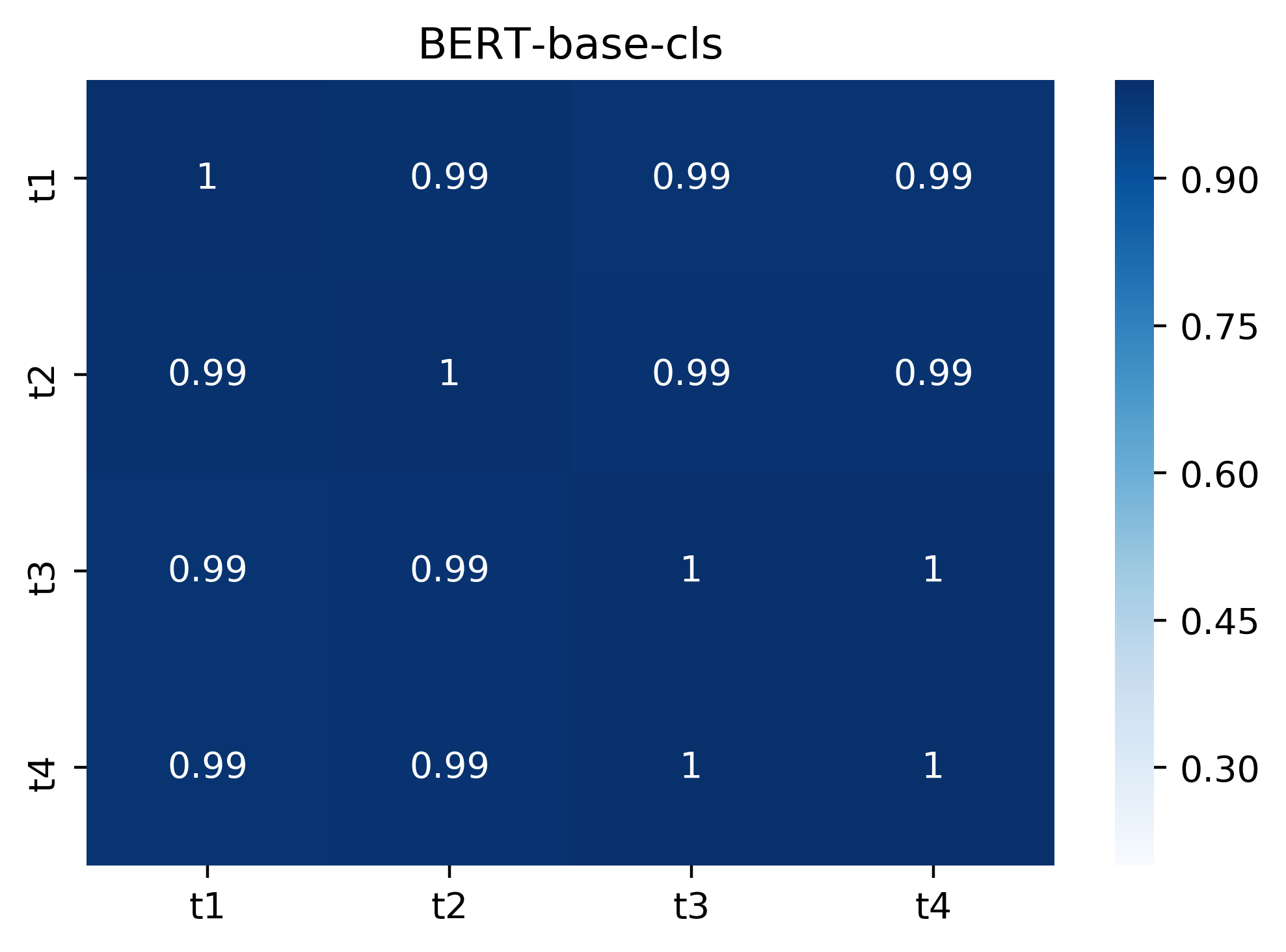}\quad
\includegraphics[width=0.3\textwidth]{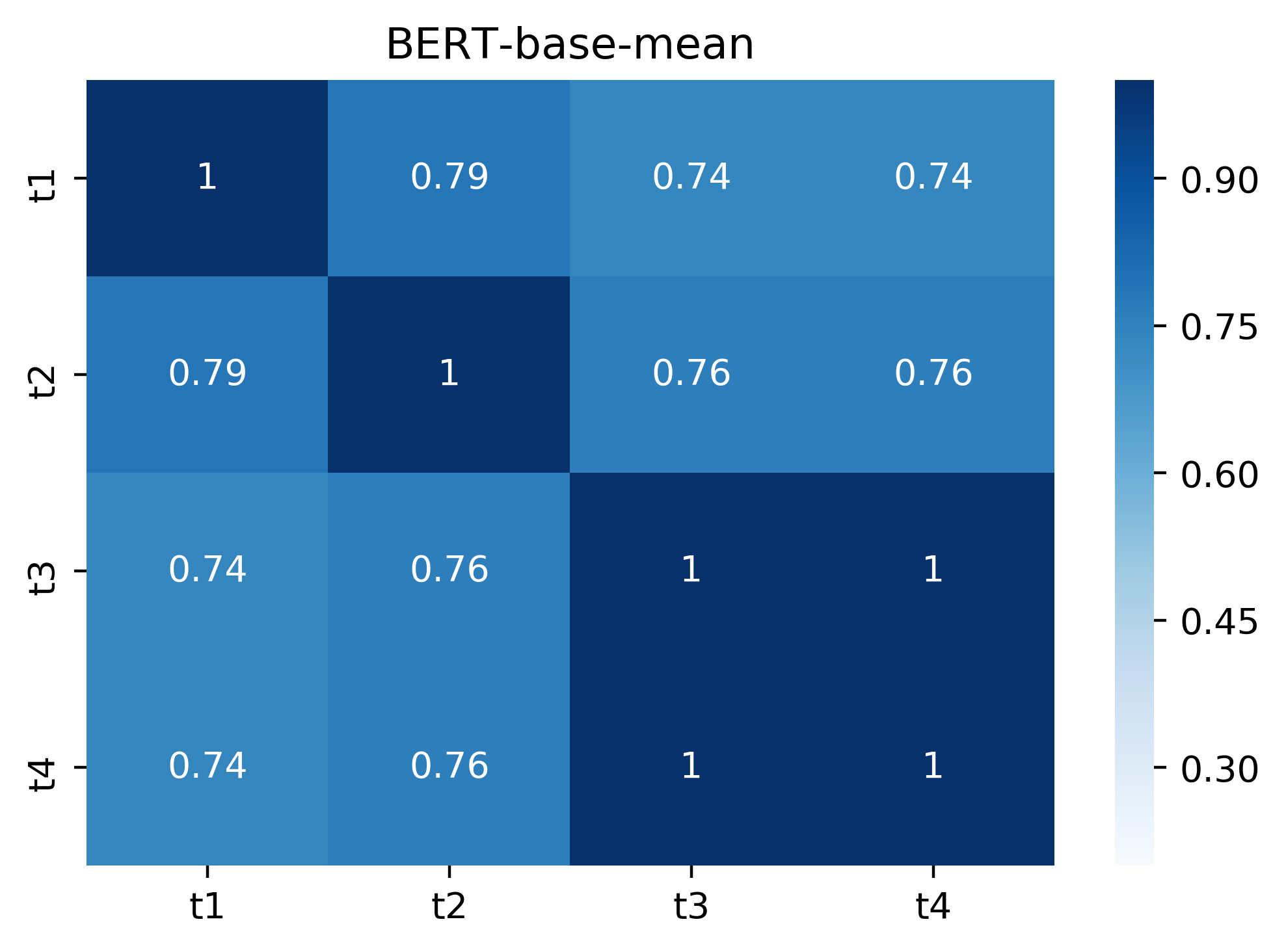}\quad
\includegraphics[width=0.3\textwidth]{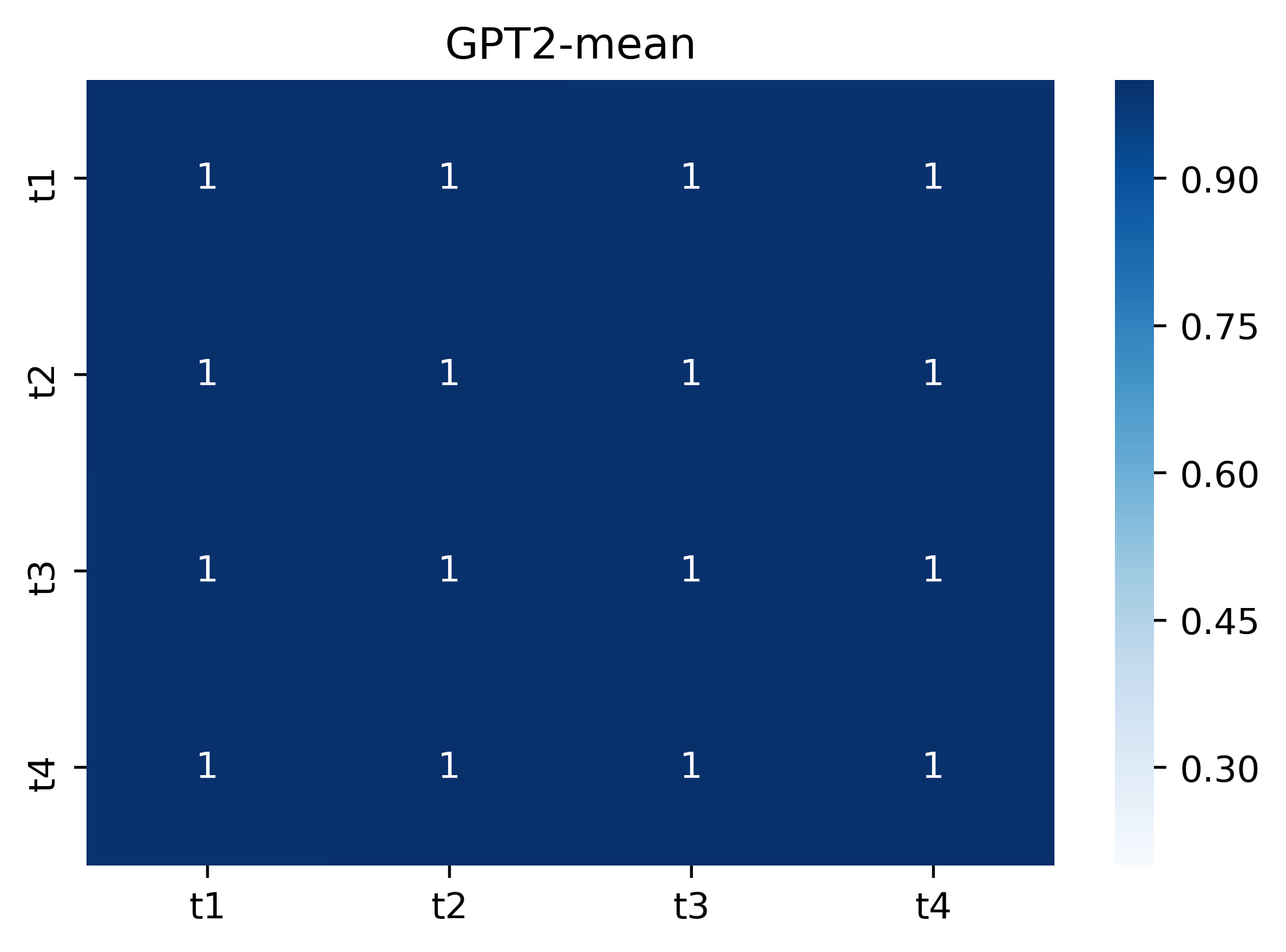}

\medskip

\includegraphics[width=0.3\textwidth]{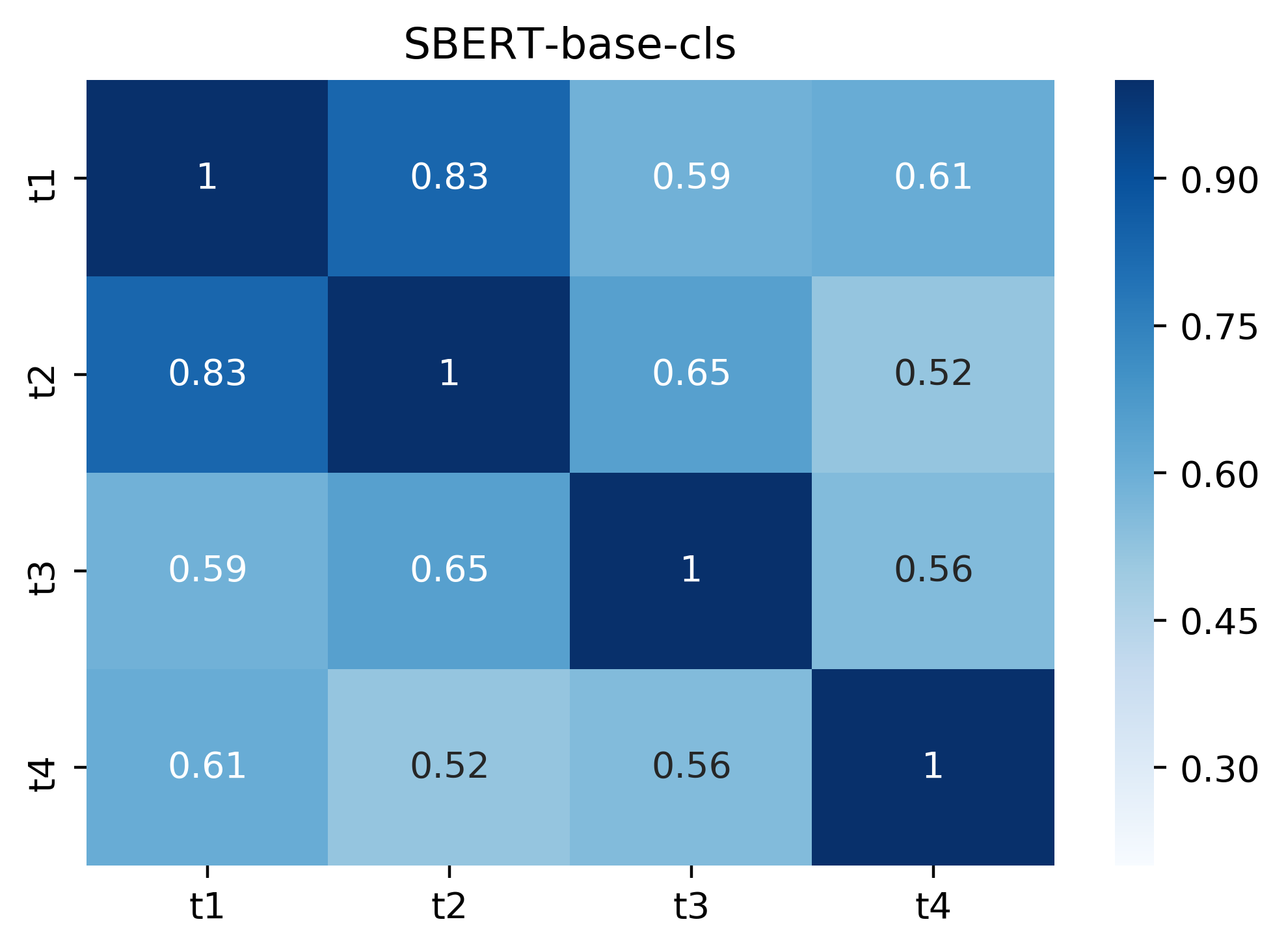}\quad
\includegraphics[width=0.3\textwidth]{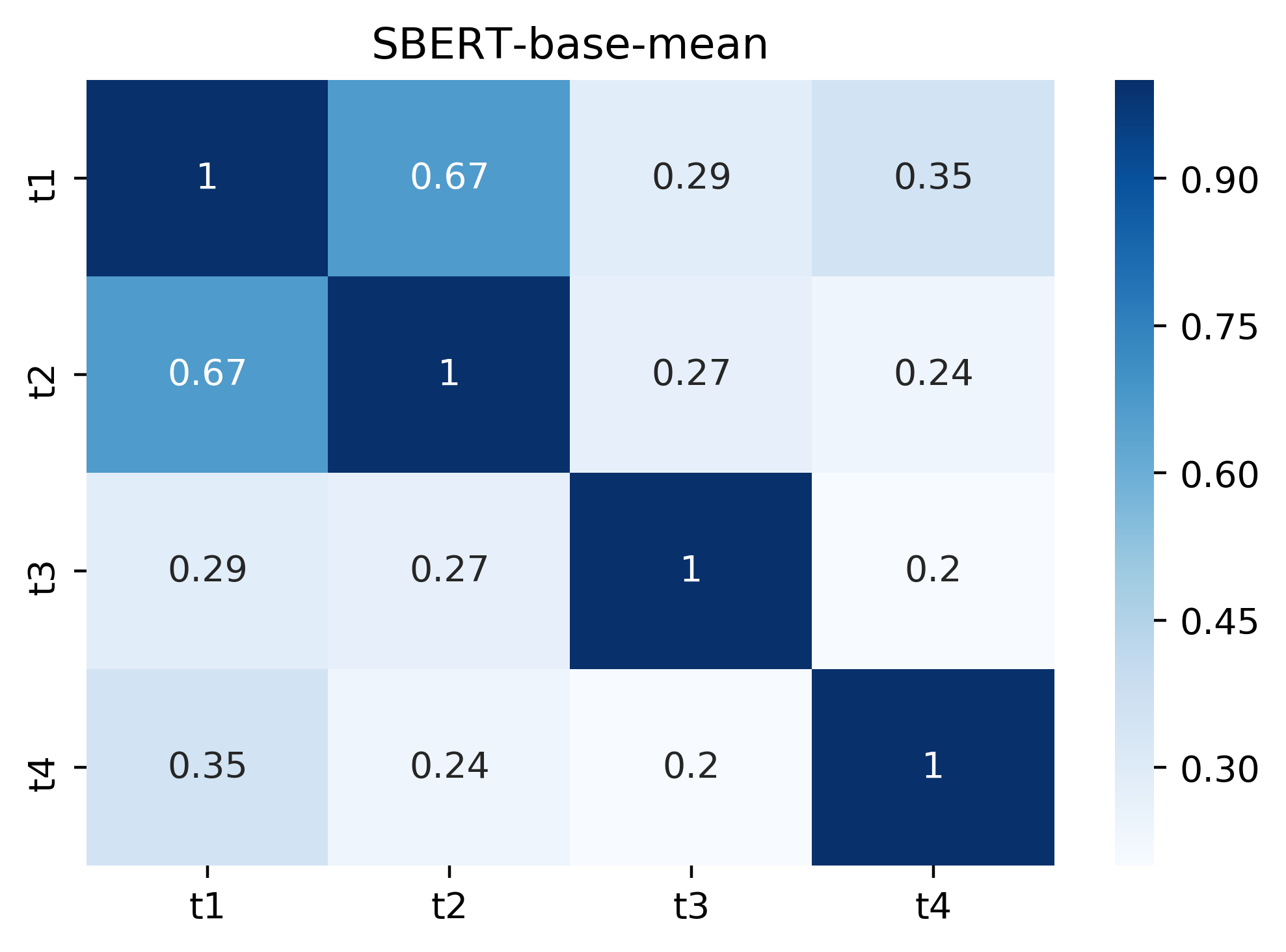}

\caption{Pairwise cosine similarity heatmap for four question sentences, where \(t1\): "How much does it cost?", \(t2\): "What is the price for subscription?", \(t3\): "What's the name of the company?", \(t4\): "What's the total height of the vehicle?"}
\label{fig2}
\end{figure*}

\section{Experiment}
\subsection{Dataset}
Our data are customer support and sales call transcripts. A customer support call is typically an interaction in which a customer calls into a support centre with a problem or question relating to a particular product or service offered by the company, and then a customer service representative of the company tries to understand and solve the problem to the customer's satisfaction. A sales call is typically an interaction in which a sales representative calls a client or potential client to discuss a potential sale of a product, or subscription to a product, either to negotiate pricing or to discuss the client’s product-specific needs (or some combination of both).

From these transcripts we automatically identified and extracted customer-side questions. Note that these questions are part of a real, human conversation, and thus are mostly short in length, and use context-dependent devices such as pronouns. For instance, under the topic of “Pricing”, typical questions include examples like “How much does it cost?” or “What’s the price for monthly subscription?” In addition, as our transcripts are produced by our ASR engine directly from speech, some sentences may contain transcription errors. 

The questions that occur naturally in human conversation are very different from the questions found in public NLP data sets such as SQuAD. The SQuAD questions test reading comprehension. The questions can be interpreted without prior discourse context.

Table 2 shows a comparison between SQuAD and our naturally occurring questions found in our data. The SQuAD questions are longer compared to our data. However, the percentage of stop words in our data indicates that questions asked in human conversations tend to use more stop words than those in SQuAD.

\begin{table*}
\centering
\begin{tabular}{| c| c| c | c|}
 &
Prec@50 &
Prec@100 &
Prec@200 \\
BERT-base-cls &
0.16 &
0.09 &
0.05 \\
BERT-base-mean &
0.74 &
0.55 &
0.37 \\
GPT2-mean &
0.04 &
0.04 &
0.03 \\
SBERT-base-cls &
0.92 &
0.88 &
0.79 \\
SBERT-base-mean &
0.98 &
0.90 &
0.78 \\
\end{tabular}
%}
\caption{Semantic matching precision with unweighted score at three cutoff points: 50, 100 and 200. 
}
\label{table3}
\end{table*}

\begin{table*}[t]
\centering
\begin{tabular}{| p{3.5cm}| c| c | c| c| c|}
 &
Prec@50 &
Prec@100 &
Prec@150 &
Prec@200 &
HR@Prec=0.9 \\
SBERT-base-mean with unweighted score &
0.98 &
0.90 &
0.86 &
0.78 &
0.0100 \\
SBERT-base-mean with weighted score &
0.98 &
0.93 &
0.87 &
0.79 &
0.0123 \\
\end{tabular}
%}
\caption{Comparison of weighted and unweighted similarity score with SBERT-base-mean. Weighted score slightly outperforms unweighted ones at cutoff points.}
\label{table4}
\end{table*}

\subsection{Similarity Score Distribution}
To better understand the distribution of sentence embeddings produced by each contextualized model and how they perform when directly used in similarity evaluation, we conducted the following experiments to test the similarity score distribution for each model: we sampled 1,000 question sentences from our dataset, and calculated the cosine similarity for all pairs of embeddings.

Figure 1 shows the calculated pairwise question cosine similarity distributions for different contextualized models. The similarity scores from BERT and GPT-2 follow a narrow distribution, i.e. any two questions have high cosine similarity, whether they are semantically related or not. Use of the BERT \textit{CLS} token or average pooling of the last hidden layer as the sentence representations does not greatly affect the pairwise similarity spread. In contrast, the SBERT scores exhibit a much wider distribution. Recall that SBERT was fine-tuned on natural language inference tasks, which is clearly beneficial for the task of determining the semantic similarity of questions.

\citep{ethayarajh-2019-contextual} refers to the phenomenon of very strong cosine similarity between vector representations as \textit{anisotropy}. The author shows that anisotropy is found in word-level representations across all layers in many contextualized models including BERT and GPT-2. Our experiment further shows that it also applies to sentence-level representations in BERT and GPT-2. Anisotropy can be problematic when using embeddings directly for sentence similarity comparison tasks, as all sentences are similar in some senses. As shown in Figure 2, we tested the pairwise cosine similarity of four different question sentences. Only \(t1\) and \(t2\) should have relatively high score as they are semantically close, but BERT and GPT-2 produced similar high scores for all pairs that cannot be distinguished easily. While SBERT successfully differentiated pairs according to their semantic relevance.

\subsection{Evaluation}
\subsubsection{Selecting Exemplars}
In this experiment, we seeded a question topic “Pricing” for the semantic matching. Following the procedures described in section 3.2, we selected ten questions related to “Pricing” as our exemplars. As a practical matter, ten exemplars is a reasonable number to ask a domain specialist to provide for a new topic.  These exemplar questions are all semantically similar but differ in syntactic formulation, such as “How much does it cost?” and “What’s the price for the monthly subscription?”. We use embeddings from the testing models as sentence representations and conduct the semantic matching on 10,000 sampled question sentences from our dataset described in section 4.1.

\subsubsection{Metrics}
Section 3.2 proposes two score computation methods for semantic matching, we can therefore rank the samples by the score using either approach. Since our dataset is unlabelled, our internal raters manually evaluated the precision at certain cutoff points after ranking: top 50, top 100 and top 200 samples with highest similarity score. This metric is noted as  \textbf{Prec@50} to describe the precision at top 50 samples in below experiments.

Because we do not have labels for our data, we are not able to calculate recall. Instead we calculate the “hit rate” defined as:

\begin{equation}
\label{eq:3}
HR = \frac{number\;of\;predicted\;positive\;samples}{number\;of\;all\;samples}
\end{equation}

To compare the hit rate of different approaches, we set a target precision score and calculate the hit rate at that precision level. \textbf{HR@Prec=0.9} denotes the hit rate of the level when the precision is 0.9

\begin{figure*}[t]
\centering
\includegraphics[width=1\textwidth]{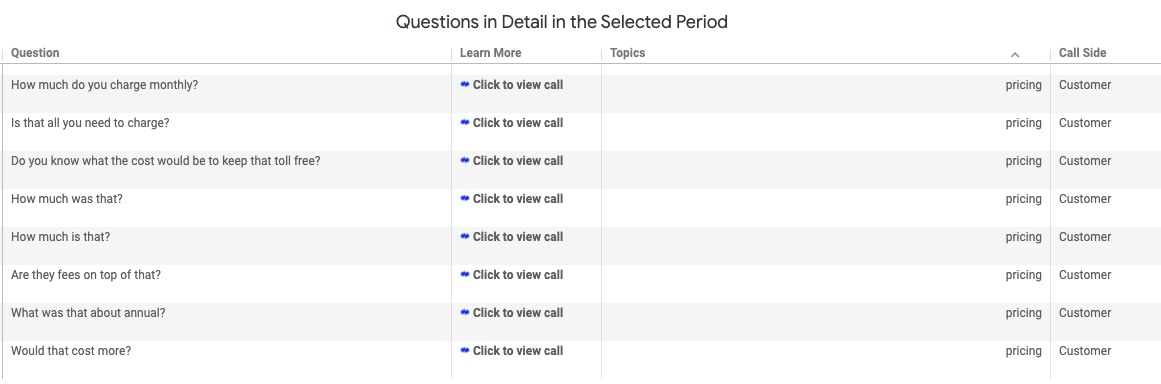} % Reduce the figure size so that it is slightly narrower than the column.
\caption{“Pricing” questions in our dashboard.}
\label{fig3}
\end{figure*}

\subsubsection{Performance Comparison}
In Table 3, we evaluate unweighted semantic matching using (\ref{eq:1}) for different embedding models. It is clear that SBERT, a BERT variant that was fine-tuned on inferences tasks, performs better at measuring the semantic similarity of sentences. Also, for BERT and SBERT, using mean as sentence representation outperforms \textit{CLS} token in this task. However, sentence representations from GPT-2 performs poorly thus are not a good fit for semantic comparison.

In discussions with business users, we have observed that when reading reports, false positives are highly distracting. We therefore require a precision score of 0.9 or higher. As Table 4 shows, the weighted score using (\ref{eq:2}) has slightly higher precision at all cutoff points, it also provides a 23\% relative improvement in hit rate over the unweighted score. This results in a higher density of true positives in the analytics reports that collate the semantically similar questions.

\section{Early-Stage Deployment}
We have implemented a data visualization dashboard (Figure 3) based on the best semantic matching approach described above (SBERT-mean with weighted similarity score). The dashboard groups semantically similar questions for topics of interest to managers such as “Pricing” and “Contact Information”. The dashboard is currently available for users enrolled in an early adopter program.

The dashboard helps the call center businesses collect and group questions asked by the customers. More importantly, it provides managers with information that they could use to coach agents (e.g. by drilling down to review the occurrence of specific questions in transcribed conversations) so that they can better prepare answers for the commonly asked questions in calls.

\section{Conclusion}
We showed that when evaluating using a directional similarity metric, sentence-level representations produced by out-of-the-box contextualized embedding models such as BERT and GPT-2 are anisotropic, which makes them unsuitable for semantic similarity comparison tasks such as grouping semantically similar questions found in natural, human-to-human conversations. However, SBERT, which fine-tunes BERT model on NLI and STS datasets mitigates the anisotropy issue, thereby improving the performance of the semantic comparisons. Combining the best sentence embedding representation with the weighted scoring functions yields more precise groupings of semantically similar sets of the naturally-occurring questions in our dataset.

The business solution that we have described only requires a few exemplar question sentences as seeds, to yield performance that meets our user needs. The solution does not require expensive data labelling to train a supervised classifier. We have implemented a visualization dashboard using this solution for our early enrolled users.

\bibliographystyle{acl_natbib}
\bibliography{custom}

% \appendix

% \section{Example Appendix}
% \label{sec:appendix}

% This is an appendix.

\end{document}